\documentclass[11pt,a4paper]{article}
\usepackage[hyperref]{acl2021}
\usepackage{times}
\usepackage{latexsym}

\usepackage{microtype}

\aclfinalcopy % Uncomment this line for the final submission
\usepackage{url}
\usepackage{booktabs}       % professional-quality tables
\usepackage{amsfonts}       % blackboard math symbols
\usepackage{nicefrac}       % compact symbols for 1/2, etc.
\usepackage{microtype}      % microtypography

\usepackage{times}

\usepackage{eqnarray,amsbsy,amsmath,amssymb,amsthm}
\usepackage{algorithm}
\usepackage{graphicx}

\usepackage{algorithmic}
\usepackage{subcaption}
\usepackage{multirow}

\usepackage{tikz}
\usepackage{csquotes}
\usetikzlibrary{bayesnet} 
\usepackage{bbm}

\usepackage{csquotes}
\usepackage{mathtools}
\usepackage{comment}
\usepackage{natbib}
\usepackage{enumitem}
\usepackage{xspace} 

\graphicspath{{pictures/}}

\newcommand{\bx}{\mathbf{x}}

\newcommand{\by}{\mathbf{y}}

\newcommand{\old}[1]{}
\newcommand{\eat}[1]{}

\usepackage{soul}
\usepackage{color}
\definecolor{aquamarine}{rgb}{0.5, 1.0, 0.83}
\definecolor{Mycolor2}{HTML}{00F9DE}

\usepackage{bm}
\usepackage{multirow}
\usepackage{tikz}
\usetikzlibrary{positioning,chains,fit,shapes,calc}
\usepackage{pgfplots}

\definecolor{turquoise}{cmyk}{.43,0,.24,0} 
\definecolor{myblue}{RGB}{80,80,160}
\definecolor{mygreen}{RGB}{143,255,204}

\tikzstyle{plate caption} = [caption, node distance=0, inner sep=-0.pt,below left=1pt and 0pt of #1.south east]
\tikzstyle{plate} = [draw, rectangle, fit=#1]

\graphicspath{{pictures/}}

\usetikzlibrary{calc}
\usepackage{zref-savepos}
\usepackage{tabu}

\title{Encouraging Neural Machine Translation to Satisfy \\ Terminology Constraints}

\author{Melissa Ailem, Jinghsu Liu and Raheel Qader   \\
Lingua Custodia, France\\ 
{\tt \{melissa.ailem,jingshu.liu,raheel.qader\}@linguacustodia.com}  \\}

\date{}

\begin{document}
\maketitle
\begin{abstract}
We present a new approach to encourage neural machine translation to satisfy lexical constraints. Our method acts at the training step and thereby avoiding the introduction of any extra computational overhead at inference step. The proposed method combines three main ingredients. The first one consists in augmenting the training data to specify the constraints. Intuitively, this encourages the model to learn a copy behavior when it encounters constraint terms. Compared to previous work, we use a simplified augmentation strategy without source factors. The second ingredient is constraint token masking, which makes it even easier for the model to learn the copy behavior and generalize better. The third one, is a modification of the standard cross entropy loss to bias the model towards assigning high probabilities to constraint words. Empirical results show that our method improves upon related baselines in terms of both BLEU score and the percentage of generated constraint terms.
\end{abstract}

% !TEX root = acl2020.tex
\section{Introduction}
\label{intro}

Neural Machine Translation (NMT) systems enjoy high performance and efficient inference \citep{sutskever2014sequence, bahdanau2014neural, luong2015effective, vaswani2017attention}. However, when it comes to domain specific scenarios, where it is often necessary to take into account terminology constraints, NMT models suffer from the lack of explicit source-target correspondences making it challenging to enforce such constraints.
For instance, consider the following sentence from the financial domain : \emph{``\textbf{Holders} may submit instructions based on a minimum quantity being accepted by the \textbf{offeror.}''}. According to the financial terminology, the words \emph{Holders} and \emph{offeror} should be translated \emph{porteurs} and \emph{initiateur} respectively. Unfortunately, a generic English-French NMT model would translate the above sentence as: \emph{``Les \textbf{titulaires} peuvent soumettre des instructions en fonction d'une quantité minimale acceptée par l'\textbf{offrant}.''}, where the words \emph{Holders} and \emph{offeror} are translated into \emph{titulaires} and \emph{offrant} respectively. To address this limitation various approaches have been proposed. They can be grouped into two categories based on whether they enforce constraints at inference or at training time. The former family of methods changes the decoding step to inject the constraint terms in the output. While effective at satisfying constraints, these techniques tend to suffer from several weaknesses such as high computational cost at the decoding stage, decreased translation quality due to strict enforcement of terminology constraints \citep{hokamp2017lexically, post2018fast}, or ineptness if there are multiple constraints in the input/output \citep{susanto2020lexically}.

The other category of methods, which we follow in this work, integrates lexical constraints during training \cite{dinu2019training}. More precisely, they augment the training data in such a way as to inform the NMT model of the constrains that need to be satisfied \citep{crego2016systran, song2019code, dinu2019training}. This type of approaches has the advantage of not changing the NMT model as well as of not introducing any additional computational overheads at inference time. 
One limitation of these methods is their soft nature, i.e, not all constraints are guaranteed to be present in the output.

In this paper we pursue the latter line of research and improve upon the recent work of \citet{dinu2019training} by (i) only using tags --without source factors -- to distinguish between constraints and other words, (ii) performing constraint token masking for robustness/generalization purposes and (iii) modifying the standard cross-entropy loss to bias the model towards generating constraint terms. Empirical results show that our approaches improve both the BLEU sore and the number of satisfied constrains compared to previous work.

% !TEX root = acl2020.tex
\section{Related Work}
\label{sRelated}
Existing approaches can be cataloged based on whether they integrate constraints at inference/decoding \citep{chatterjee2017guiding, hasler2018neural, hokamp2017lexically} or at training time \citep{dinu2019training}.

Among methods of the first category, we can mention the Grid Beam Search (GBS) algorithm, which consists in reorganizing the vanilla beam search to place constraints correctly in the output as well as infer accurately the constraints-free parts.  
While successful in placing constrains compared to the original BS algorithm, GBS suffers from a high decoding time, it increases inference complexity exponentially with the number of constraints.     
To alleviate this issue, several improvement have been proposed, such as Dynamic Beam Allocation (DBA) \citep{post2018fast} and its optimized extension, namely vectorized DBA \citep{hu2019improved}. Despite an important gain in computational time, these methods still significantly increase the decoding time. For instance, the method of \citet{post2018fast} is three times slower than the constraint-free beam search. More recently, \citet{susanto2020lexically} rely on the levenstein transformer \citep{gu2019levenshtein}, which uses an edit-based decoder iteratively refining the output using deletion and insertion operations. To enforce constraints using this model, \citet{susanto2020lexically} add one step to the decoder that consists in placing constraint terms in the output, and they further disallow the deletion operation on constraint terms. Albeit effective, the main limitation of this approach is in constraint ordering -- when there is more than one constraint term in the output. That is, the initial order in which constraints have been placed remains unchanged. 

Different from the above, the second family of methods integrates lexical constraints at training time. For instance, \citet{crego2016systran} replace the terminology terms with placeholders during training and then add them back in a post-processing step. \citet{song2019code} proposed to annotate the training set by adding the target side of the terminology terms in the source sentences. A transformer model \citep{vaswani2017attention} is then trained on this augmented training set. 
This training data annotation has been also explored to teach the NMT to use translation memories \citep{gu2018search} or to enforce copy behavior \citep{pham2018towards}.
\citet{dinu2019training} proposed two different ways to augment the training data, namely the append and the replace approaches.  The former is similar to approach proposed in \citep{song2019code}, and the second requires to replace the source term of the constraints in the source sentence by its corresponding target side in the terminology entries. This method further uses source factors in order to distinguish the constraints from the rest of the source sentence. This is the closest approach to ours. The key differences are as follows. Our method uses only tags (without source factors) to specify constraints in the training set, and we further perform constraint-token masking, which improves model robustness/generalization as supported by our experiments. Moreover, we investigate a biased cross-entropy loss to encourage the NMT model to assign higher probabilities to constraint words.

% !TEX root = acl2020.tex
\section{Method}
\label{our_approach}
Our objective is to encourage neural machine translation to satisfy lexical constraints. To this end we introduce three 
%\RQ{two or three}
changes to the standard procedure, namely training data augmentation, token masking, and cross-entropy loss modification. 

%\subsection{TrAining Data Augmentation (TADA) and token MASKing (MASK)}
%\RQ{be careful, it is a Figure not table}
\paragraph{TrAining Data Augmentation (TADA).} Similar to previous work, the key idea is to bias the NMT model to exhibit a copy behavior when it encounters constraints. To this end, given some source sentence along with some constraints, we use tags to specify the constraints in the source sentence where relevant, as depicted in Figure \ref{ex1}. Note that as opposed to previous work, we do not introduce any further information (e.g., source factors), the constraints are specified using tags only.

\paragraph{Token MASKing (MASK).} We further consider masking the source part of the constraint -- tokens in blue -- as illustrated in Figure \ref{ex1} last row. We postulate that this might be useful from at least two perspectives. For one, this provides a more general pattern for the model to learn to perform the copy operation every time it encounters the tag $<S>$ followed by the MASK token. For another, this makes the model more apt to support conflicting constraints, i.e., constraints sharing the same source part but which have different target parts. This may be useful if some tokens must be translated into different targets for some specific documents and contexts at test time. 
%the mask could also help to improve the generalization of the model ... may be to be explored empirically like varying the percentage of constraints in the training data
%\subsection{Mask}
{\setlength\tabcolsep{2pt}
\def\arraystretch{1.4}  
\begin{figure}[h!]
    \centering
    \footnotesize
    \begin{tabular}{p{1.5cm}|p{6cm}}
    Source & His critics state that this will just increase the \textcolor{blue}{\textbf{budgetary deficit}} . \\
        Constraint &  \textcolor{blue}{\textbf{budgetary deficit}} $\to$  \textcolor{orange}{\textbf{Haushaltsdefizit}} \\
        \hline
       TADA  &  His critics state that this will just increase the   $<$S$>$ \textcolor{blue}{\textbf{budgetary deficit}} $<$C$>$ \textcolor{orange}{\textbf{Haushaltsdefizit}} $<$/C$>$ .\\
        +MASK & His critics state that this will just increase the $<$S$>$ \textbf{MASK MASK} $<$C$>$ \textcolor{orange}{\textbf{Haushaltsdefizit}} $<$/C$>$ .  \\
    \end{tabular}
    \caption{Illustration of TrAining Data Augmentation (TADA) and MASK. }
    \label{ex1}
\end{figure}
}
%\subsection{Weighted Cross-Entropy (WCE) Loss}
\paragraph{Weighted Cross-Entropy (WCE) Loss.}
Let $\bx = (x_1,\ldots,x_{T_x})$ denote a sentence in some input language represented as a sequence of $T_x$ words, and $\by = (y_1,\ldots,y_{T_y})$ its translation in some target language. From a probabilistic perspective 
%\paragraph{Neural Machine Translation (NMT)}  is a function that maps an input text written in a given source language to its translation into a target language.
neural machine translation can be cast as estimating the conditional probability $p(\by|\bx)$ parametrized with neural networks, and which is usually assumed to factorize as follows,
\begin{equation}
p(\by|\bx) = p(y_1 | \bx) \prod_{t=2}^{T_y} p(y_t | \bx, y_{1:t-1})
\end{equation}
where $y_{1:t-1}$ denote previously generated tokens. A predominant loss function in this context is the well know cross-entropy given by,
\begin{eqnarray}
\mathcal{L}= - \log p(\by|\bx) = - \sum_{t=1}^{T_y} \log p(y_t | \bx, y_{1:t-1})
\end{eqnarray}
As our objective is to encourage the NMT model towards generating the desired constraints, we propose to modify the above loss 
%\RQ{put as follows at the end of the sentence}
to provide a stronger learning signal to the model when it assigns a low probability to a constrain token $y_t$, as follows. 
%\paragraph{Constrained NMT} 
%In order to even more enforce the constraints in the generated translations, ....
\begin{equation}
\mathcal{L}= - \sum_{t=1}^{T} w_{y_t}  \log p(y_t | \bx, y_{1:t-1}) 
\end{equation}
where, $w_{y_t}=\alpha\geq 1$ if $y_t$ is a constraint word, and $w_{y_t}=1$ otherwise.
%\begin{equation}
% w_{y_t}=\begin{cases}
%   \alpha\geq 1 & \text{if $y_t$ is a constraint word.}\\
%   1 & \text{Otherwise}. 
%  \end{cases} \nonumber
%\end{equation}
As long as $\alpha$ is strictly greater than $1$, the model would be biased towards assigning higher probabilities to constraint tokens. In practice one can set $\alpha$ to either a fixed value (e.g., selected based on some validation set) or using some annealing heuristic, i.e., start with $\alpha=1$ and then gradually increase its value as learning progresses.

% !TEX root = acl2020.tex
\section{Experiments}

\subsection{Parallel Data}

Following previous work \citep{dinu2019training, susanto2020lexically}, we assess our approach using the WMT 2018 English-German news translation tasks\footnote{http://www.statmt.org/wmt18/translation-task.html}. Our training dataset consists of nearly 2.2 million  English-German parallel sentences from Europarl and news commentary.
To compare our approach against existing works, we use two parallel English-German test sets extracted from WMT newstest 2017, and made available by  \citet{dinu2019training} (see section \ref{terms} for details).
Following the same authors, we use WMT newstest 2013 for validation containing 3000 parallel sentences.

\vspace{-0.2cm}
\subsection{Terminologies}
\label{terms}
In order to take into account lexical constraints, training, test and validation sets were annotated using two English-German bilingual terminologies extracted from IATE\footnote{https://iate.europa.eu}  and Wiktionary\footnote{https://www.wiktionary.org/}.
The two test sets released by \citep{dinu2019training} have been extracted from WMT 2017 using IATE and Wiktionary respectively. The lexical constraints are added in the source sentences when source and target terms in the dictionaries entries are present in source and target sentences in the parallel dataset respectively. The test set extracted using IATE (wiktionary) contains 414 (727) sentences and 452 (884) term annotations.
The training and validation sets have been annotated using both dictionaries making sure there is no overlap with the term annotations used in the test sets.
For the training dataset, only 10\% of the original data have been annotated with lexical constraints in order to preserve as far as possible the same performance when the model is not terminology-grounded \citep{dinu2019training}.

{\def\arraystretch{1.5}    
\begin{figure*}[!t]
    \footnotesize
    \centering
    \begin{tabular}{p{2cm}|p{13cm}}
    \hline
    \multicolumn{2}{c}{Without MASK} \\
    \hline
      Source & For a while, one major problem has been finding homes subsequently for refugees that have been given $<$S$>$ \textbf{certified} $<$C$>$ \textcolor{blue}{\textbf{anerkannt}} $<$/C$>$ status. \\
       TADA  & Seit geraumer Zeit besteht ein großes Problem darin, Häuser für Flüchtlinge zu finden, die  \textcolor{orange}{\textbf{zertifiziert}} wurden.  \\
       \hline 
        \multicolumn{2}{c}{With MASK} \\
        \hline 
        Source &  For a while, one major problem has been finding homes subsequently for refugees that have been given $<$S$>$ \textbf{MASK} $<$C$>$ \textcolor{blue}{\textbf{anerkannt}} $<$/C$>$ status.\\
          TADA+MASK & Seit einiger Zeit besteht ein großes Problem darin, später Heime für Flüchtlinge zu finden , die  \textcolor{blue}{\textbf{anerkannt}} wurden.  \\
          + WCE Loss & 	Seit einiger Zeit besteht ein großes Problem darin, später Heime für Flüchtlinge zu finden, die \textcolor{blue}{\textbf{anerkannt}} worden sind. \\
          \hline
          \hline 
          Target & Ein schwerwiegendes Problem ist es seit einiger Zeit , Wohnungen für die Anschlussflüchtlinge zu finden, die \textcolor{blue}{\textbf{anerkannt}} worden sind.   \\
          \hline
    \end{tabular}
    % \vspace{-0.1cm}
    \caption{ IATE : Example of en-de translation generated with TADA only and with TADA+MASK. With TADA only we observe that a variant of the target side of the constraint has been used (zertifiziert). In contrast, with MASK we observe that the target side of the constraint has been copied directly. Furthermore, using WCE loss  leads to a translation which is even closer to the ground truth. }
    \label{qual1}
\end{figure*}}

{\def\arraystretch{1.5}   
\begin{figure*}[!t]
    \footnotesize
    \centering
    \begin{tabular}{p{2cm}|p{13cm}}
    \hline
    \multicolumn{2}{c}{Without MASK} \\
    \hline
      Source &  If perpetrators have to leave the country quicker , that will boost security and increase the $<$S$>$ \textbf{general public} $<$C$>$  \textcolor{blue}{\textbf{Bevölkerung}} $<$/C$>$ 's $<$S$>$ \textbf{approval} $<$C$>$  \textcolor{blue}{\textbf{Zustimmung}} $<$/C$>$ of refugee politics .\\
       TADA  &  Wenn die Täter das Land schneller verlassen müssen , wird dies die Sicherheit erhöhen und die \textcolor{blue}{\textbf{Zustimmung}} der Öffentlichkeit zur Flüchtlingspolitik erhöhen .\\
       \hline 
        \multicolumn{2}{c}{With MASK} \\
        \hline 
        Source & If perpetrators have to leave the country quicker , that will boost security and increase the $<$S$>$ \textbf{MASK MASK} $<$C$>$  \textcolor{blue}{\textbf{Bevölkerung}} $<$/C$>$ 's $<$S$>$ \textbf{MASK} $<$C$>$  \textcolor{blue}{\textbf{Zustimmung}} $<$/C$>$ of refugee politics .  \\
          TADA+MASK & Wenn die Täter das Land schneller verlassen müssen , wird dies die Sicherheit erhöhen und die  \textcolor{blue}{\textbf{Zustimmung}} der  \textcolor{blue}{\textbf{Bevölkerung}} zur Flüchtlingspolitik erhöhen . \\
          + WCE Loss & Wenn die Täter das Land schneller verlassen müssen , wird dies die Sicherheit erhöhen und die  \textcolor{blue}{\textbf{Zustimmung}} der  \textcolor{blue}{\textbf{Bevölkerung}} zur Flüchtlingspolitik erhöhen . 	 \\
          \hline
          \hline 
          Target &    Wenn Straftäter schneller das Land verlassen müssten , erhöhe das aber die Sicherheit und stärke auch die \textcolor{blue}{\textbf{Zustimmung}} der \textcolor{blue}{\textbf{Bevölkerung}} für die Flüchtlingspolitik .\\
          \hline
    \end{tabular}
    % \vspace{-0.1cm}
    \caption{ IATE : Example with multiple constraints. With TADA  we observe that only one constraint is satisfied. Adding MASK makes it possible to satisfy both constraints.   }
    \label{qual2}
\end{figure*}}

%\vspace{-0.6cm}
\subsection{Settings}

We use Moses tokenizer \citep{koehn2007moses} to tokenize our corpus and we learn a joint source and target BPE encoding  \citep{sennrich2015neural} with 40k merge operations to segment it into sub-word units, resulting in a vocabulary size of 40388 words.  
Our models are trained using the transformer architecture  \citep{vaswani2017attention} with three stacked encoders and decoders.
The same hyperparameters as in \citep{dinu2019training} were used where source  and target embeddings are tied with the softmax layer. The models are trained for a minimum of 50 epochs and a maximum of 100 epochs with a batch size of 3000 tokens per iteration. Our validation set WMT 2013 is used to compute the stopping criterion. We use a beam size of 5 during inference for all models.
Regarding the proposed WCE Loss, we start training with $\alpha=1$ for the first ninety epochs, then we continue learning for ten more epochs with $\alpha=2$. In a pilot experiment, we explored different strategies to set the value of $\alpha$, such as using $\alpha>1$ from the beginning of training, increase the value of $\alpha$ every 5/10 iterations by $+0.1$, or train with $\alpha=1$ for most iterations and then set $\alpha$ to a higher value (e.g., $\alpha=2$) for the last few iterations. We retained the latter approach as it worked best among the ones we investigated. 
%\footnote{In a pilot experiments, we explored different strategies to set the value of alpha, such as using $\alpha>1$ from the beginning of training or train with alpha=1 for most iterations and then set alpha to a higher value $\alpha=2$  for the last few iterations. The latter approach worked best among the ones we investigated.}.

% We then selected the best epoch that minimizes the most our WCE Loss. 
{\setlength\tabcolsep{4pt}
\def\arraystretch{1.2}      
\begin{table}[h!]
    \centering
    \footnotesize
    \begin{tabular}{c|c|c|c|c}
         & \multicolumn{2}{c|}{IATE}&  \multicolumn{2}{c}{Wiktionary} \\
         \hline
         & Term\%& BLEU  & Term\% & BLEU\\
         \hline
         \multicolumn{5}{l}{\emph{Previous works}} \\
         \hline
         Transformer$^\dagger$ & 76.30 & 25.80& 76.90 & 26.00  \\
         Const. Dec.$^\ddagger$ & 82.00 &25.30 &99.50 & 25.80  \\
         Source. Fact.$^\S$ & 94.50 &26.00 & 93.40 & 26.30 \\
         \hline 
           \multicolumn{5}{l}{\emph{Our work}} \\
           \hline
            %TADA+MASK &96.43 &25.33 &93.96 & 25.30 \\
           TADA+MASK & 97.80  &26.89  & 96.55& 26.69 \\
            +WCE Loss & \textbf{98.02} & \textbf{27.11} &  \textbf{96.84} & \textbf{26.73} \\
            \hline
    \end{tabular}
    \scriptsize{ $^\dagger$:\citep{vaswani2017attention},  $^\ddagger$: \citep{post2018fast}, $^\S$: \citep{dinu2019training} }
            \caption{Comparison with baselines in terms of BLEU score and Term usage percentage.}
    \label{res}
\end{table}}

\begin{figure*}[h!]
    \centering
    \begin{subfigure}{0.7\columnwidth}
    \includegraphics[scale=0.25]{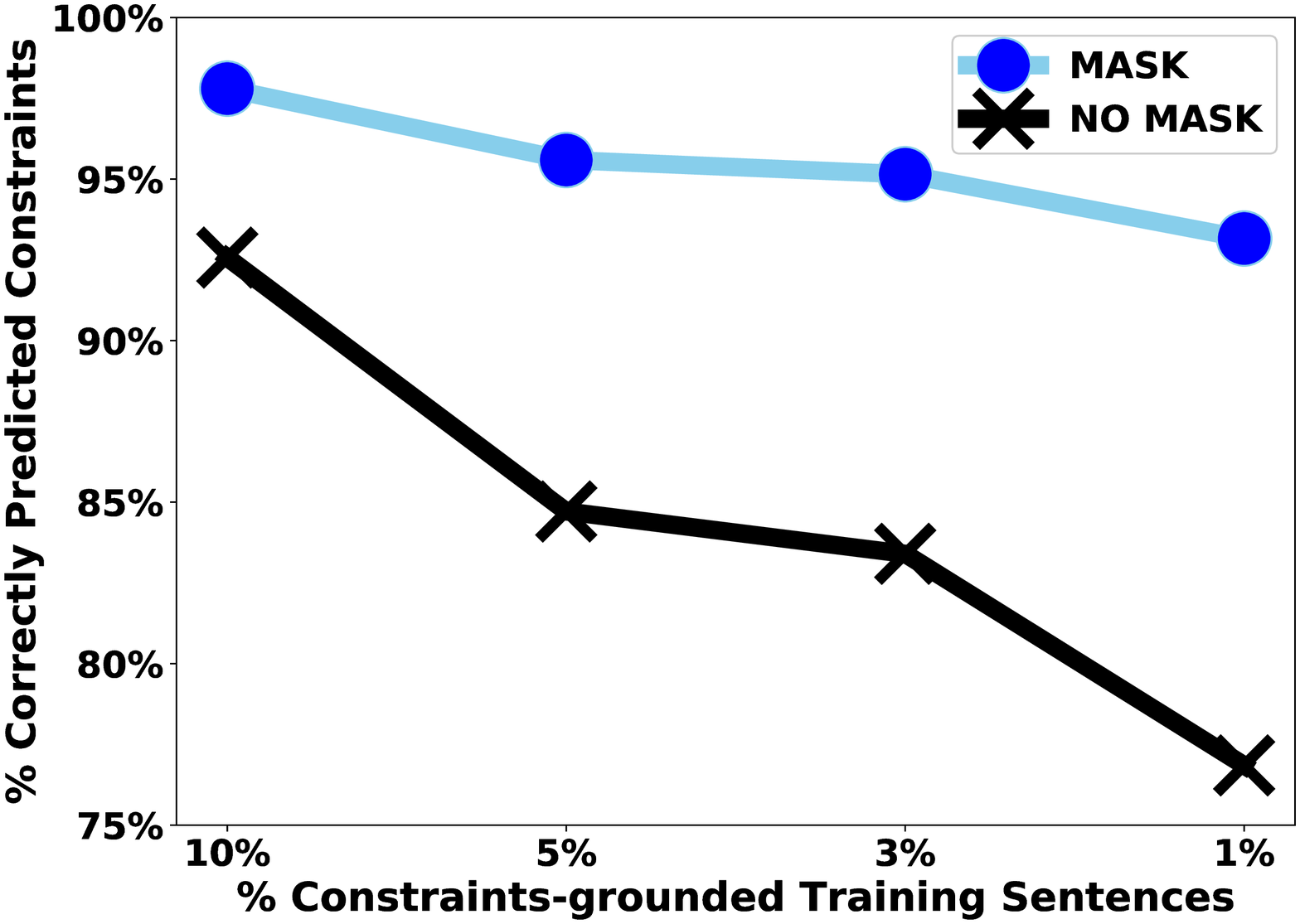}
    \subcaption{\scriptsize IATE test set}
    \end{subfigure}
    \hspace{1.3cm}
    \begin{subfigure}{0.7\columnwidth}
    \includegraphics[scale=0.25]{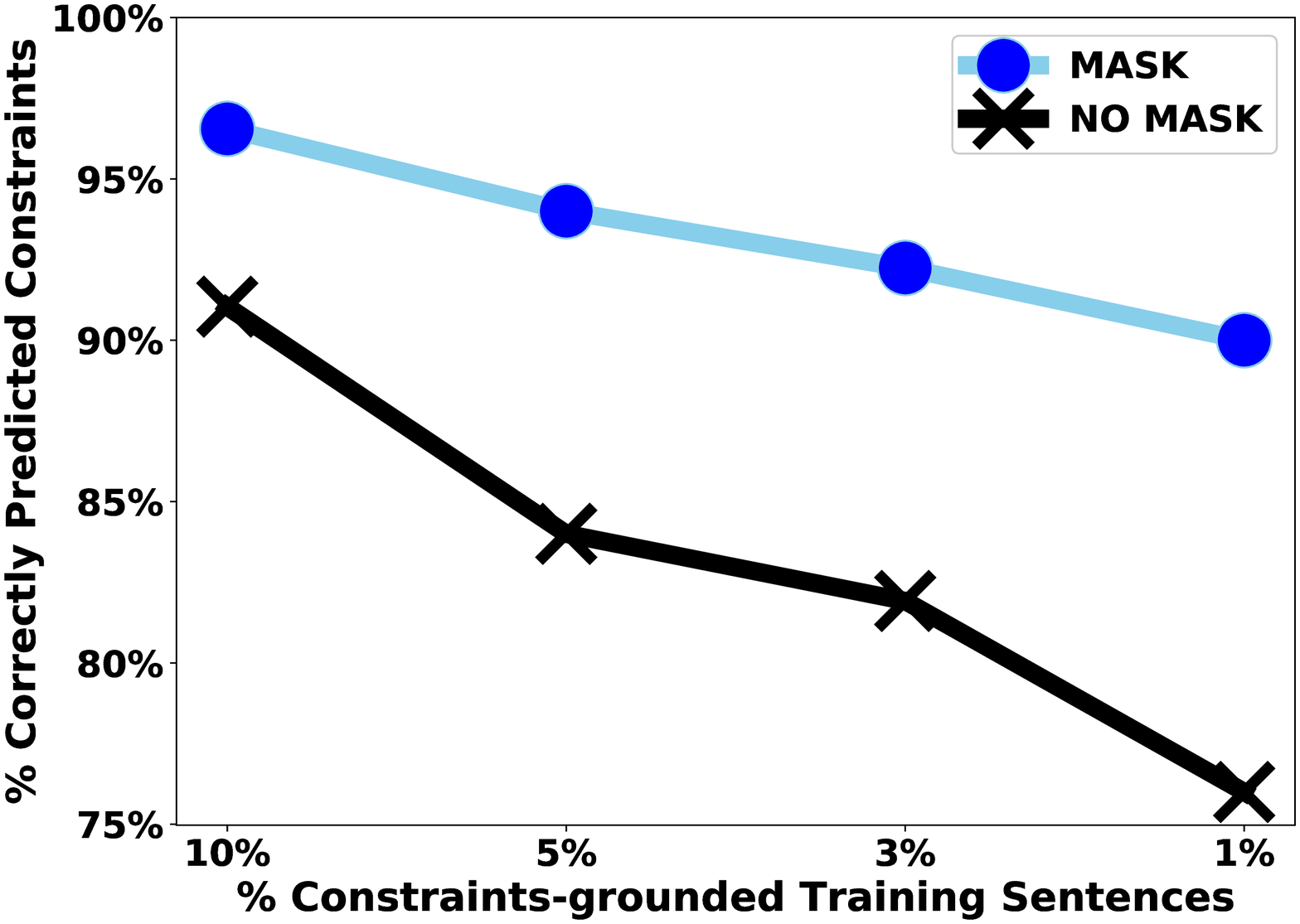}
    \subcaption{\scriptsize Wiktionary test set}
    \end{subfigure}
    \caption{{Percentage of correctly generated constraints with 10\%, 5\%, 3\% and 1\% of constraint-grounded training source sentences. With MASK the NMT model is less sensitive to the diminution of the percentage constraint-grounded sentences.}    }
    \label{fig_ablation}
%\vspace{-0.2in}
\end{figure*}

\subsection{Results}
We compare our approach to related NMT models integrating terminology constraints in terms of BLEU score \citep{papineni2002bleu} and term usage rate (Term\%),
which is defined as the ratio between the number of constraints generated by the model and the total number of constraints. 
%=======est ce que cest parreil que ca ? number of constraints generated in the output divided by the total number of the given constraints (phrase de levenstein).=========
The results are presented in Table \ref{res}, and the main findings are as follows.

{\bf Comparison with baselines.} Our methods significantly outperform the baselines in terms of both the BLEU score and the percentage of correctly generated constraint terms.
%We observe that using only the Training Data Augmentation (TADA) and the MASK, both BLEU score and the percentage of constraints present in the predicted translations are significantly improved.
TADA+MASK increases the BLEU score with +0.89\% and +0.39\% for IATE and Wiktionary test sets respectively. Regarding constraints (Terms\%) we observe an improvement of +3.3\%  for IATE. Using the WCE loss further improves performances. For Wiktionary, Constrained Decoder reaches the highest terminology-use rate. However, the latter method suffers from a high decoding time and decreases translation quality. %, as reported in previous work as well \citep{dinu2019training}.

{\bf Importance of MASK.} To assess the impact of token masking, we report in Figure \ref{fig_ablation} the performance of TADA and TADA+MASK when the percentage of constraint annotations used in the training varies from 10\% to 1\%. Using MASK makes the model more robust to the diminution of the percentage of constraint-grounded sentences.
The qualitative examples of Figures \ref{qual1} and \ref{qual2} further illustrate the benefit of token masking. In the former example, masking the source part of the constraint ``certified'' seems to have prevented the model from generating ``zertifiziert'' -- see Figure \ref{qual1}'s caption for details. Figure \ref{qual2} shows a translation example containing multiple constraints to be satisfied. It seems that the use of MASK makes the model more apt to effectively handle and satisfy all the constraints. This is not necessarily the case of the model without MASK, which satisfies one constraint only. The results of Figures \ref{qual1}, \ref{qual2} and \ref{fig_ablation} provide empirical support for the benefits of the proposed token masking in model generalization and robustness.   

{\bf Impact of the WCE loss.} To assess the impact of the WCE loss, we revisit the results of Table \ref{res} and the examples of Figures \ref{qual1} and \ref{qual2}. In all cases, we observe that using the proposed weighted cross-entropy loss further improves the quality of translation and the percentage of generated constraints, which demonstrate the benefits of biasing the model towards generating constraints tokens.

\label{expe}

\section{Conclusion}

%In this paper, we propose a new constrained training approach to integrate lexical constraints in domain specific settings. Our approach augment the training set and constrain the loss function to force the NMT model to take into account the constraints.

To encourage neural machine translation to satisfy terminology constraints, we propose an approach combining training data augmentation and token masking with a weighted cross-entropy loss. Our method is architecture independent and in principle it can be applied to any NMT model. Experiments on real-world datasets show that the proposed approach improves upon recent related baselines in terms of both BLEU score and the percentage of generated constraint terms.

In face of the multiplicity of methods to integrate terminology constraints, an interesting future work is to consider combining our method with other techniques within an ensemble approach.

\section{Acknowledgments}
This work was partially funded by the French Ministry of Defense. 

\bibliographystyle{acl_natbib}
\bibliography{acl2021}

\begin{thebibliography}{19}
\expandafter\ifx\csname natexlab\endcsname\relax\def\natexlab#1{#1}\fi

\bibitem[{Bahdanau et~al.(2014)Bahdanau, Cho, and Bengio}]{bahdanau2014neural}
Dzmitry Bahdanau, Kyunghyun Cho, and Yoshua Bengio. 2014.
\newblock Neural machine translation by jointly learning to align and
  translate.
\newblock \emph{arXiv preprint arXiv:1409.0473}.

\bibitem[{Chatterjee et~al.(2017)Chatterjee, Negri, Turchi, Federico, Specia,
  and Blain}]{chatterjee2017guiding}
Rajen Chatterjee, Matteo Negri, Marco Turchi, Marcello Federico, Lucia Specia,
  and Fr{\'e}d{\'e}ric Blain. 2017.
\newblock Guiding neural machine translation decoding with external knowledge.
\newblock In \emph{Proceedings of the Second Conference on Machine
  Translation}, pages 157--168.

\bibitem[{Crego et~al.(2016)Crego, Kim, Klein, Rebollo, Yang, Senellart,
  Akhanov, Brunelle, Coquard, Deng et~al.}]{crego2016systran}
Josep Crego, Jungi Kim, Guillaume Klein, Anabel Rebollo, Kathy Yang, Jean
  Senellart, Egor Akhanov, Patrice Brunelle, Aurelien Coquard, Yongchao Deng,
  et~al. 2016.
\newblock Systran's pure neural machine translation systems.
\newblock \emph{arXiv preprint arXiv:1610.05540}.

\bibitem[{Dinu et~al.(2019)Dinu, Mathur, Federico, and
  Al-Onaizan}]{dinu2019training}
Georgiana Dinu, Prashant Mathur, Marcello Federico, and Yaser Al-Onaizan. 2019.
\newblock Training neural machine translation to apply terminology constraints.
\newblock \emph{Proceedings of the 57th Annual Meeting of the Association for
  Computational Linguistics}, page 3063–3068.

\bibitem[{Gu et~al.(2019)Gu, Wang, and Zhao}]{gu2019levenshtein}
Jiatao Gu, Changhan Wang, and Junbo Zhao. 2019.
\newblock Levenshtein transformer.
\newblock In \emph{Advances in Neural Information Processing Systems}, pages
  11181--11191.

\bibitem[{Gu et~al.(2018)Gu, Wang, Cho, and Li}]{gu2018search}
Jiatao Gu, Yong Wang, Kyunghyun Cho, and Victor~OK Li. 2018.
\newblock Search engine guided neural machine translation.
\newblock In \emph{Proceedings of the AAAI Conference on Artificial
  Intelligence}, volume~32.

\bibitem[{Hasler et~al.(2018)Hasler, De~Gispert, Iglesias, and
  Byrne}]{hasler2018neural}
Eva Hasler, Adri{\`a} De~Gispert, Gonzalo Iglesias, and Bill Byrne. 2018.
\newblock Neural machine translation decoding with terminology constraints.
\newblock \emph{Proceedings of NAACL-HLT}, page 506–512.

\bibitem[{Hokamp and Liu(2017)}]{hokamp2017lexically}
Chris Hokamp and Qun Liu. 2017.
\newblock Lexically constrained decoding for sequence generation using grid
  beam search.
\newblock \emph{Proceedings of the 55th Annual Meeting of the Association for
  Computational Linguistics}, page 1535–1546.

\bibitem[{Hu et~al.(2019)Hu, Khayrallah, Culkin, Xia, Chen, Post, and
  Van~Durme}]{hu2019improved}
J~Edward Hu, Huda Khayrallah, Ryan Culkin, Patrick Xia, Tongfei Chen, Matt
  Post, and Benjamin Van~Durme. 2019.
\newblock Improved lexically constrained decoding for translation and
  monolingual rewriting.
\newblock In \emph{Proceedings of the 2019 Conference of the North American
  Chapter of the Association for Computational Linguistics: Human Language
  Technologies, Volume 1 (Long and Short Papers)}, pages 839--850.

\bibitem[{Koehn et~al.(2007)Koehn, Hoang, Birch, Callison-Burch, Federico,
  Bertoldi, Cowan, Shen, Moran, Zens et~al.}]{koehn2007moses}
Philipp Koehn, Hieu Hoang, Alexandra Birch, Chris Callison-Burch, Marcello
  Federico, Nicola Bertoldi, Brooke Cowan, Wade Shen, Christine Moran, Richard
  Zens, et~al. 2007.
\newblock Moses: Open source toolkit for statistical machine translation.
\newblock In \emph{Proceedings of the 45th annual meeting of the ACL on
  interactive poster and demonstration sessions}, pages 177--180. Association
  for Computational Linguistics.

\bibitem[{Luong et~al.(2015)Luong, Pham, and Manning}]{luong2015effective}
Minh-Thang Luong, Hieu Pham, and Christopher~D Manning. 2015.
\newblock Effective approaches to attention-based neural machine translation.
\newblock \emph{Proceedings of the 2015 Conference on Empirical Methods in
  Natural Language Processing}, page 1412–1421.

\bibitem[{Papineni et~al.(2002)Papineni, Roukos, Ward, and
  Zhu}]{papineni2002bleu}
Kishore Papineni, Salim Roukos, Todd Ward, and Wei-Jing Zhu. 2002.
\newblock Bleu: a method for automatic evaluation of machine translation.
\newblock In \emph{Proceedings of the 40th annual meeting of the Association
  for Computational Linguistics}, pages 311--318.

\bibitem[{Pham et~al.(2018)Pham, Niehues, and Waibel}]{pham2018towards}
Ngoc-Quan Pham, Jan Niehues, and Alex Waibel. 2018.
\newblock Towards one-shot learning for rare-word translation with external
  experts.
\newblock \emph{Proceedings of the 2nd Workshop on Neural Machine Translation
  and Generation}, page 100–109.

\bibitem[{Post and Vilar(2018)}]{post2018fast}
Matt Post and David Vilar. 2018.
\newblock Fast lexically constrained decoding with dynamic beam allocation for
  neural machine translation.
\newblock \emph{Proceedings of NAACL-HLT 2018}, page 1314–1324.

\bibitem[{Sennrich et~al.(2015)Sennrich, Haddow, and
  Birch}]{sennrich2015neural}
Rico Sennrich, Barry Haddow, and Alexandra Birch. 2015.
\newblock Neural machine translation of rare words with subword units.
\newblock \emph{Proceedings of the 54th Annual Meeting of the Association for
  Computational Linguistics}, page 1715–1725.

\bibitem[{Song et~al.(2019)Song, Zhang, Yu, Luo, Wang, and
  Zhang}]{song2019code}
Kai Song, Yue Zhang, Heng Yu, Weihua Luo, Kun Wang, and Min Zhang. 2019.
\newblock Code-switching for enhancing nmt with pre-specified translation.
\newblock \emph{Proceedings of the 2019 Conference of the North American
  Chapter of the Association for Computational Linguistics: Human Language
  Technologies, Volume 1 (Long and Short Papers)}, page 449–459.

\bibitem[{Susanto et~al.(2020)Susanto, Chollampatt, and
  Tan}]{susanto2020lexically}
Raymond~Hendy Susanto, Shamil Chollampatt, and Liling Tan. 2020.
\newblock Lexically constrained neural machine translation with levenshtein
  transformer.
\newblock \emph{Proceedings of the 58th Annual Meeting of the Association for
  Computational Linguistics}, page 3536–3543.

\bibitem[{Sutskever et~al.(2014)Sutskever, Vinyals, and
  Le}]{sutskever2014sequence}
Ilya Sutskever, Oriol Vinyals, and Quoc~V Le. 2014.
\newblock Sequence to sequence learning with neural networks.
\newblock \emph{arXiv preprint arXiv:1409.3215}.

\bibitem[{Vaswani et~al.(2017)Vaswani, Shazeer, Parmar, Uszkoreit, Jones,
  Gomez, Kaiser, and Polosukhin}]{vaswani2017attention}
Ashish Vaswani, Noam Shazeer, Niki Parmar, Jakob Uszkoreit, Llion Jones,
  Aidan~N Gomez, {\L}ukasz Kaiser, and Illia Polosukhin. 2017.
\newblock Attention is all you need.
\newblock In \emph{Advances in neural information processing systems}, pages
  5998--6008.

\end{thebibliography}

%\appendix

\end{document}